\documentclass[11pt,a4paper]{article}
\usepackage[hyperref]{acl2020}
\usepackage{times}
\usepackage{latexsym}
\usepackage{textcomp}

\usepackage{amsmath,amsfonts,bm}

\def\eqref#1{equation~\ref{#1}}

\def\1{\bm{1}}

\DeclareMathAlphabet{\mathsfit}{\encodingdefault}{\sfdefault}{m}{sl}
\SetMathAlphabet{\mathsfit}{bold}{\encodingdefault}{\sfdefault}{bx}{n}

\newcommand{\softmax}{\mathrm{softmax}}

\DeclareMathOperator*{\argmax}{arg\,max}
\DeclareMathOperator*{\argmin}{arg\,min}

\usepackage{microtype}

\usepackage{amsmath}
\usepackage{amssymb}
\usepackage{graphicx}
\usepackage{subcaption}
\usepackage{multirow}
\usepackage{xspace}

\usepackage{booktabs}
\usepackage{adjustbox}
\usepackage{supertabular}

\usepackage{makecell}

\newcommand{\x}{\boldsymbol{x}}
\newcommand{\y}{\boldsymbol{y}}
\newcommand{\z}{\boldsymbol{z}}

\newcommand{\q}{\boldsymbol{q}}

\newcommand{\onehot}{\textit{onehot}}
\newcommand{\loss}{\textit{Loss}}

\newcommand{\infnet}{\mathbf{A}_{\Psi}}

\newcommand{\oone}{\mathbf{O_1}}
\newcommand{\otwo}{\mathbf{O_2}}
\newcommand{\ogen}{\mathbf{O}}
\newcommand{\name}{ENGINE\xspace}

\aclfinalcopy 

\definecolor{realpurple}{RGB}{87, 6, 140}

\newcommand{\given}{\mid}

\title{ENGINE: Energy-Based Inference Networks for \\ Non-Autoregressive Machine Translation}

\author{Lifu Tu$^1$ \ \ \ \ \ \ \ Richard Yuanzhe Pang$^2$\thanks{~\ Work partly done at Toyota Technological Institute at Chicago and the University of Chicago.} \ \ \ \ \ \ \ Sam Wiseman$^1$ \ \ \ \ \ \ \ Kevin Gimpel$^1$ \\
$^1$Toyota Technological Institute at Chicago, Chicago, IL 60637, USA \\
$^2$New York University, New York, NY 10011, USA \\
{\tt \{lifu,swiseman,kgimpel\}@ttic.edu, yzpang@nyu.edu}}

\date{}

\begin{document}
\maketitle
\begin{abstract}
We propose to train a non-autoregressive machine translation model to minimize the energy defined by a pretrained autoregressive model. In particular, we view our non-autoregressive translation system as an inference network \citep{tu-18} trained to minimize the autoregressive 
teacher energy. 
This contrasts with the popular approach of training a non-autoregressive model on a distilled corpus consisting of the beam-searched outputs of such a teacher model. 
Our approach, which we call \name (ENerGy-based Inference NEtworks),  
achieves state-of-the-art non-autoregressive 
results on the IWSLT 2014 DE-EN and WMT 2016 RO-EN datasets, approaching the performance of 
autoregressive models.\footnote{Code 
is available at \url{https://github.com/lifu-tu/ENGINE}}
\end{abstract}

\section{Introduction}

The performance of non-autoregressive neural machine translation (NAT) systems, which predict tokens in the target language independently of each other conditioned on the source sentence, has been improving steadily in recent  years~\citep{lee-etal-2018-deterministic,ghazvininejad-etal-2019-mask,ma-etal-2019-flowseq}. One common ingredient in getting non-autoregressive systems to perform well is to train them on a corpus of distilled translations~\citep{DBLP:conf/emnlp/KimR16}. This distilled corpus consists of source sentences paired with the translations produced by a pretrained autoregressive ``teacher'' system.

As an alternative to training non-autoregressive translation systems on distilled corpora, 
we instead propose to train them to minimize the \textit{energy} defined by a pretrained  autoregressive teacher model. That is, we view non-autoregressive machine translation systems as inference networks~\citep{tu-18,tu-gimpel-2019-benchmarking,tu2019improving} trained to minimize the teacher's energy. 
This provides the non-autoregressive model with additional information related to the energy of the teacher, rather than just the approximate minimizers of the teacher's energy appearing in a distilled corpus. 

In order to train inference networks to minimize an energy function, the energy must be differentiable with respect to the inference network output. 
We describe several approaches for relaxing the autoregressive teacher's energy to make it amenable to minimization with an inference network, and compare them empirically. 
We experiment with two non-autoregressive inference network architectures, one based on bidirectional RNNs and the other based on the transformer model of \citet{ghazvininejad-etal-2019-mask}. 

In experiments on the IWSLT 2014 DE-EN and WMT 2016 RO-EN datasets, we show that 
training to minimize the teacher's energy significantly outperforms training with distilled outputs. 
Our approach, which we call \name (ENerGy-based Inference NEtworks), achieves state-of-the-art results for non-autoregressive translation on these datasets, approaching the results of the 
autoregressive teachers. 
Our hope is that \name will enable energy-based models to be applied more broadly for non-autoregressive generation in the future.

\section{Related Work}

Non-autoregressive neural machine translation began with the work of \citet{gu2018non}, who found benefit from using knowledge distillation~\citep{hinton2015distilling}, and in particular sequence-level distilled outputs~\citep{DBLP:conf/emnlp/KimR16}. 
Subsequent work has narrowed the gap between non-autoregressive and autoregressive translation, including multi-iteration refinements~\citep{lee-etal-2018-deterministic, ghazvininejad-etal-2019-mask,saharia2020nonautoregressive,kasai2020parallel} and rescoring with autoregressive models~\citep{kaiser2018fast,wei-etal-2019-imitation,ma-etal-2019-flowseq,NIPS2019_8566}.  \citet{ghazvininejad2020aligned} and \citet{saharia2020nonautoregressive} proposed aligned cross entropy or latent alignment models and achieved the best results of all non-autoregressive models without refinement or rescoring. 
We propose training inference networks with autoregressive energies and outperform the best purely non-autoregressive methods.

Another related approach trains an ``actor'' network to manipulate the hidden state of an autoregressive neural MT system~\citep{gu-etal-2017-trainable,chen-etal-2018-stable,zhou20iclr} in order to bias it toward outputs with better BLEU scores. This work modifies the original pretrained network rather than using it to define an energy for training an inference network. 

Energy-based models have had limited application in text generation due to the computational challenges involved in learning and inference in extremely large search spaces~\citep{bakhtin2020energybased}. The use of inference networks to output approximate minimizers of a loss function is popular in variational inference~\citep{kingma2013autoencoding, rezende2014stochastic}, and, more recently, in structured prediction~\citep{tu-18,tu-gimpel-2019-benchmarking,tu2019improving}, including previously for neural MT~\citep{DBLP:conf/aaai/GuIL18}.

\section{Energy-Based Inference Networks for Non-Autoregressive NMT}

Most neural machine translation (NMT) systems model the conditional distribution $p_{\Theta}(\y \given \x)$ of a target sequence $\y = \langle y_1, y_2,..., y_{T}\rangle$ given a source sequence $\x = \langle x_1, x_2,..., x_{T_s}\rangle$, where each $y_t$ comes from a vocabulary $\mathcal{V}$, $y_T$ is $\langle \mathit{eos}\rangle$, and $y_{0}$ is $\langle \mathit{bos}\rangle$. 
It is common in NMT to define this conditional distribution using an ``autoregressive'' factorization~\citep{NIPS2014_5346,BahdanauCB14,NIPS2017_7181}: 
\begin{align}
    \log p_{\Theta}(\y \given \x) =  \sum_{t=1}^{|\y|} \log p_{\Theta}(y_t \mid \y_{0:t-1}, \x) \nonumber
\end{align}
This model can be viewed as an energy-based model~\citep{lecun-06} by defining the \textbf{energy function} $E_\Theta(\x, \y)= -\log p_{\Theta}(\y \given \x)$. 
Given trained parameters $\Theta$, 
test time inference seeks to find the translation for a given source sentence $\x$ with the lowest energy: $\hat{\y} = \argmin_{\y} \,E_\Theta(\x, \y)$.

\label{sec:infnet}

Finding the translation that minimizes the energy involves combinatorial search. 
In this paper, we train \textbf{inference networks} to perform this search approximately. 
The idea of this approach is to replace the test time combinatorial search typically employed in structured prediction  with the output of a network trained to produce approximately optimal predictions~\citep{tu-18,tu-gimpel-2019-benchmarking}. 
More formally, 
we define an inference network $\infnet$ which maps an input $\x$ to a translation $\y$ and 
is trained with the goal that 
$\infnet(\x) \approx \argmin_{\y} E_\Theta(\x, \y)$.

Specifically, we train the inference network parameters $\Psi$ as follows (assuming $\Theta$ is pretrained and fixed):
\begin{align}
\widehat{\Psi} = \argmin_{\Psi}\!\!\sum_{\langle\x,\y\rangle \in \mathcal{D}} \!\! E_\Theta(\x, \infnet(\x))\! \label{eq:objective} 
\end{align}
\noindent where $\mathcal{D}$ is a training set of sentence pairs. The network architecture of $\infnet$ can be different from the architectures used in the energy function. In this paper, we combine an autoregressive energy function with a non-autoregressive inference network. By doing so, we seek to combine the effectiveness of the autoregressive energy with the fast inference speed of a non-autoregressive network.

\subsection{Energies for Inference Network Training}
\label{sec:energies-infnet}

\begin{figure}[t]
\centering
\includegraphics[width=0.45\textwidth]{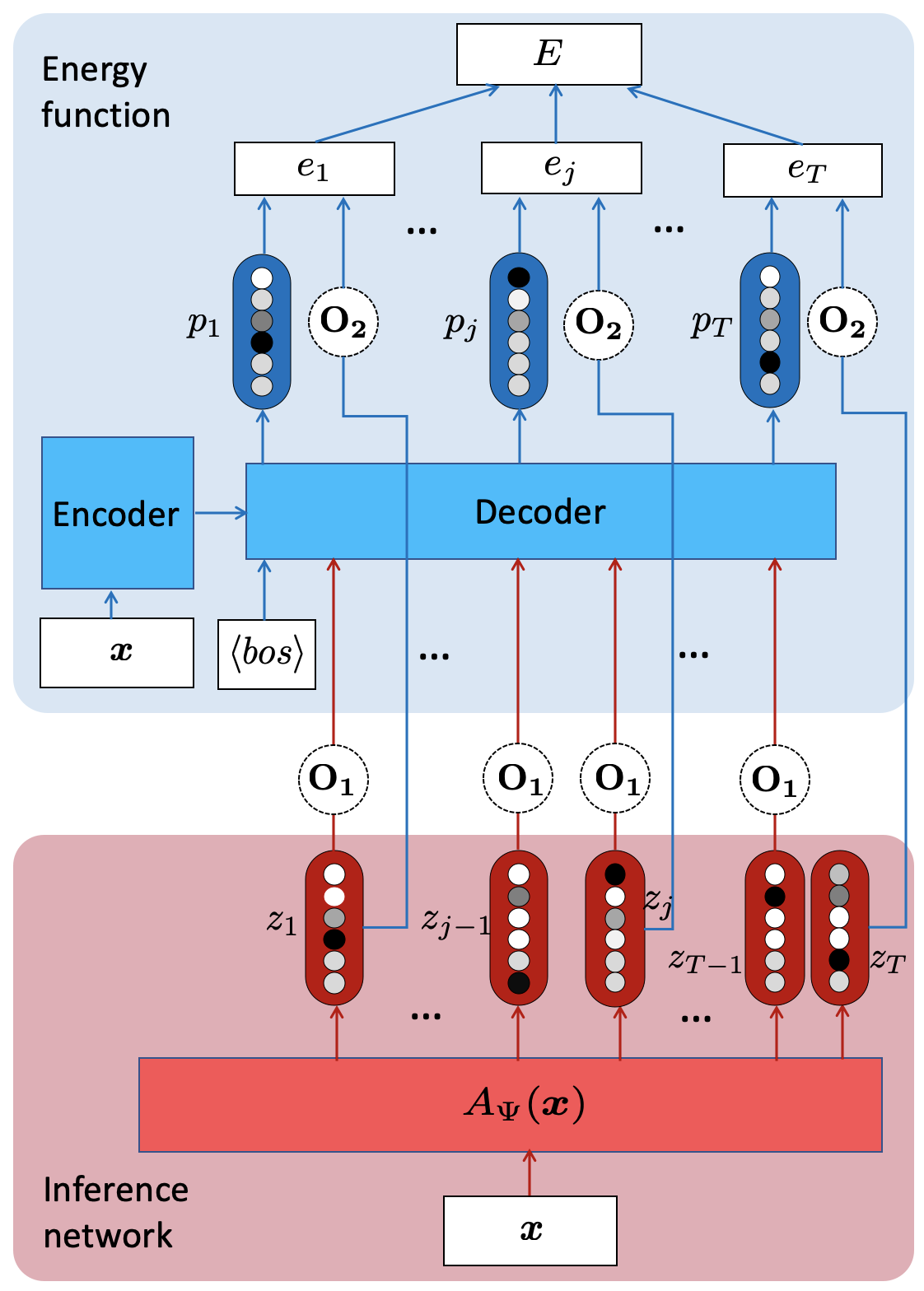}
\caption{The \name framework trains a non-autoregressive inference network $\infnet$ to produce translations with low energy under a pretrained autoregressive energy $E$. 
\label{fig:model}}
\end{figure}

In order to allow for gradient-based optimization of the inference network parameters $\Psi$, we now define a more general family of  
energy functions for NMT. First, we change the representation of the translation $\y$ in the energy, redefining $\y=\langle \y_0,\dots, \y_{|\y|} \rangle$ as a sequence of 
\textbf{\emph{distributions}} over words instead of a sequence of words.

In particular, we consider the generalized energy
\begin{equation}
E_\Theta(\x, \y) = \sum_{t=1}^{|\y|} e_t (\x,\y)
\label{eq:loss}  
\end{equation}
where
\begin{equation}
e_t(\x, \y) = - \y_t^\top \log p_{\Theta} (\cdot \mid \y_{0}, \y_{1}, \dots, \y_{t-1}, \x).
\label{eq:localen}
\end{equation}
We use the $\cdot$ notation in $p_{\Theta} (\cdot \mid \ldots)$ above to indicate that we may need the full distribution over words. Note that by replacing the $\y_t$ with one-hot distributions we recover the original energy.

In order to train an inference network to minimize this energy, we simply need a network architecture that can produce a sequence of word distributions, which is satisfied by recent non-autoregressive NMT models~\citep{ghazvininejad-etal-2019-mask}. However, because the distributions involved in the original energy are one-hot, it may be advantageous for the inference network too to output distributions that are one-hot or approximately so. We will accordingly view inference networks as producing a sequence of $T$ logit vectors $\z_t \in \mathbb{R}^\mathcal{|V|}$, and we will consider two operators $\oone$ and $\otwo$ that will be used to map these $\z_t$ logits into distributions for use in the energy. Figure~\ref{fig:model} provides an overview of our approach, including this generalized energy function, the inference network, and the two operators $\oone$ and $\otwo$. We describe choices for these operators in the next section.

\begin{table}[t]
  \begin{center}
  \small
    \begin{tabular}{lcc}
      \toprule
       & $\ogen(\z)$ & $\frac{\partial \ogen(\z)}{\partial \z}$ \\
      \midrule
      \textbf{SX} & $\q$ & $ \frac{\partial \q}{\partial \z}$ \\[1ex]
      \textbf{STL} & $\onehot{ (\argmax (\z))}$ & $\boldsymbol{I}$ \\ [1ex]
      \textbf{SG} &  $\onehot{ (\argmax (\tilde{\q}))}$ & $ \frac{\partial \tilde{\q}}{\partial \tilde{\z}}$ \\[1ex]
    \textbf{ST} & $\onehot{ (\argmax (\q))}$ & $ \frac{\partial \q}{\partial \z}$ \\[1ex]
     \textbf{GX} & $\tilde{\q}$  &  $\frac{\partial \tilde{\q}}{\partial \tilde{\z}}$ \\[1ex] 
     \bottomrule
    \end{tabular}
    \caption{Let $\ogen(\z) \, {\in} \, \Delta^{|\mathcal{V}|-1}$ be the result of applying an $\oone$ or $\otwo$ operation to logits $\z$ output by the inference network. Also let $\tilde{\z} \, {=} \, \z \, {+} \, \boldsymbol{g}$, where $\boldsymbol{g}$ is Gumbel noise, $\q \, {=} \, \softmax(\z)$, and $\tilde{\q} \, {=} \, \softmax(\tilde{\z})$. We show the Jacobian (approximation) $\frac{\partial \ogen(\z)}{\partial \z}$ we use when computing  $\frac{\partial \loss}{\partial \z} \, {=} \, \frac{\partial \loss}{\partial \ogen(\z)} \frac{\partial \ogen(\z)}{\partial \z}$, for each $\ogen(\z)$ considered. 
    }
     \label{tab:table1}
  \end{center}
\end{table}

\subsection{Choices for Operators}
\label{sec:operations}

We now consider ways of defining the two operators that govern the interface between the inference network and the energy function. 
As shown in Figure~\ref{fig:model}, we seek an operator $\oone$ to modulate the way that logits $\z_t$ output by the inference network are fed to the decoder input slots in the energy function, and an operator $\otwo$ to determine how the distribution $p_{\Theta} (\cdot \mid \ldots)$ is  used to compute the log probability of a word in $\y$. Explicitly, then, we rewrite each local energy term (Eq.~\ref{eq:localen}) as
\begin{align*} 
e_t&(\x, \y) = - \otwo(\z_t)^\top \\
&\log p_{\Theta} (\cdot \mid \nonumber \oone(\z_{0}), \oone(\z_{1}), \dots, \oone(\z_{t-1}), \x),
\end{align*}
which our inference networks will minimize with respect to the $\z_t$.

The choices we consider for $\oone$ and $\otwo$, which we present generically for operator $\ogen$ and logit vector $\z$, are shown in Table~\ref{tab:table1}, and described in more detail below. Some of these $\ogen$ operations are not differentiable, and so the Jacobian matrix $\frac{\partial \ogen(\z)}{\partial \z}$ must be approximated during learning; we show the approximations we use in Table~\ref{tab:table1} as well.

We consider five choices for each $\ogen$:

\begin{itemize}
\item[(a)] \textbf{SX}: $\softmax$. Here $\ogen(\z) \, {=} \, \softmax(\z)$; no Jacobian approximation is necessary.

\item[(b)] \textbf{STL}: straight-through logits. Here $\ogen(\z) \, {=} \, \onehot{(\argmax_i \z})$. $\frac{\partial \ogen(\z)}{\partial \z}$ is approximated by the identity matrix $\boldsymbol{I}$ (see \citet{BengioLC13}).

\item[(c)] \textbf{SG}: straight-through Gumbel-Softmax. Here $\ogen(\z) \, {=} \, \onehot{(\argmax_i \softmax(\z + \boldsymbol{g})})$, where $g_i$ is Gumbel noise.\footnote{$g_i=-\log(-\log(u_i))$ and $u_i \sim \mathbf{Uniform}(0,1)$.} $\frac{\partial \ogen(\z)}{\partial \z}$ is approximated with $\frac{\partial \, \softmax(\z + \boldsymbol{g})}{\partial \z}$ \citep{jang2016categorical}.

\item[(d)] \textbf{ST}: straight-through. This setting is identical to SG with $\boldsymbol{g} \, {=} \, \boldsymbol{0}$ (see ~\citet{BengioLC13}).

\item[(e)] \textbf{GX}: Gumbel-Softmax. Here $\ogen(\z) \, {=} \, \softmax(\z + \boldsymbol{g})$, where again $g_i$ is Gumbel noise; no Jacobian approximation is necessary.
\end{itemize}

\begin{table*}[t]
\centering
    \setlength{\tabcolsep}{2pt}
    \begin{subtable}{\columnwidth}
      \centering
      \small
        \begin{tabular}{c  ccccc}
    \toprule
	$\oone \setminus \otwo$ & SX & STL & SG & ST & GX\\ 
    \midrule
    SX & \textbf{55 (20.2)} & 256 (0) & 56 (19.6) & \textbf{55 (20.1)} & 55 (19.6) \\
    STL & 97 (14.8) & 164 (8.2)  & 94 (13.7) & 95 (14.6) & 190 (0)\\
    SG & 82 (15.2) & 206 (0) & 81 (14.7) & 82 (15.0)  & 83 (13.5)\\
    ST & 81 (14.7) & 170 (0) & 81 (14.4) & 80 (14.3) & 83 (13.7)  \\
    GX & \textbf{53 (19.8)} & 201 (0) & 56 (18.3) & 54 (19.6)  & 55 (19.4) \\
    \bottomrule
    \end{tabular}
    \caption{seq2seq AR energy, BiLSTM inference networks}
    \end{subtable}
    \  \ 
    \begin{subtable}{\columnwidth}
      \centering
      \small
        \begin{tabular}{  ccccc}
    \toprule
	SX & STL & SG & ST & GX \\ 
    \midrule
       \textbf{80 (31.7)} &  133 (27.8) &  81 (31.5) &  \textbf{80 (31.7)} & 81 (31.6)\\
      186 (25.3) & 133 (27.8)  & 95 (20.0) & 97 (30.1) & 180 (26.0)\\
    
      98 (30.1) & 133 (27.8) & 95 (30.1) & 97 (30.0) & 97 (29.8)\\

      98 (30.2) & 133 (27.8) & 95 (30.0) & 97 (30.1) & 97 (30.0)\\
       81 (31.5) & 133 (27.8) &  81 (31.2) &  81 (31.5)  & 81 (31.4) \\
    \bottomrule
    \end{tabular}
    \caption{transformer AR energy, CMLM inference networks} 
    \end{subtable} 
    \caption{Comparison of operator choices in terms of energies (BLEU scores) 
on the IWSLT14 DE-EN dev set with two energy/inference network combinations. 
Oracle lengths are used for decoding. $\oone$ is the operation for feeding inference network outputs into the decoder input slots in the energy. $\otwo$ is the operation for computing the energy on the output. Each row corresponds to the same $\oone$, and each column corresponds to the same $\otwo$.  
}
\label{tab:argmax}
\end{table*}

\section{Experimental Setup}

\subsection{Datasets} 
We evaluate our methods on two datasets: IWSLT14 German (DE) $\rightarrow$ English (EN) and WMT16 Romanian (RO) $\rightarrow$ English (EN). 
All data are tokenized and then segmented into subword units using byte-pair encoding \citep{sennrich-etal-2016-neural}. We use the data provided by \citet{lee-etal-2018-deterministic} for RO-EN. 

\subsection{Autoregressive Energies}
We consider two architectures for the pretrained autoregressive (AR) energy function. 
The first is an autoregressive sequence-to-sequence (seq2seq) model with attention~\citep{luong-etal-2015-effective}. The encoder is a two-layer BiLSTM with 512 units in each direction, the decoder is a two-layer LSTM with 768 units, and the word embedding size is 512. 
The second is an autoregressive transformer model~\citep{NIPS2017_7181}, where both the encoder and decoder have 6 layers, 8 attention heads per layer, model dimension 512, and hidden dimension 2048.

\subsection{Inference Network Architectures} 
We choose two different architectures: a BiLSTM ``tagger'' (a 2-layer BiLSTM followed by a fully-connected layer) and a conditional masked language
model (CMLM; \citealp{ghazvininejad-etal-2019-mask}), 
a transformer with 6 layers per stack, 8 attention heads per layer, model dimension 512, 
and hidden dimension 2048. Both architectures require the target sequence length in advance; methods for handling length are discussed in Sec.~\ref{sec:lengths}. 
For baselines, we train these inference network architectures as non-autoregressive models using the standard per-position cross-entropy loss. 
For faster inference network training, we initialize inference networks with the baselines trained with cross-entropy loss in our experiments. 

The baseline CMLMs use the partial masking strategy
described by \citet{ghazvininejad-etal-2019-mask}. This involves using some masked input tokens and some provided input tokens during training. At test time, multiple iterations (``refinement iterations'') can be used for improved results~\citep{ghazvininejad-etal-2019-mask}. Each iteration uses partially-masked input from the preceding iteration. We consider the use of multiple refinement iterations for both the CMLM baseline and the CMLM inference network.\footnote{The CMLM inference network is trained according to Eq.~\ref{eq:objective} with full masking (no partial masking like in the CMLM baseline). However, since the CMLM inference network is initialized using the CMLM baseline, which is trained using partial masking, the CMLM inference network is still compatible with refinement iterations at test time.}

\subsection{Hyperparameters}
For inference network training, the batch size is 1024 tokens. We train with the Adam optimizer \citep{kingma2015adam}. We tune the learning rate in $\{5\mathrm{e}\!-\!4, 1\mathrm{e}\!-\!4, 5\mathrm{e}\!-\!5, 1\mathrm{e}\!-\!5, 5\mathrm{e}\!-\!6, 1\mathrm{e}\!-\!6\}$. For regularization, we use L2 weight decay with rate 0.01, and dropout with rate 0.1. 
We train all models for 30 epochs. For the baselines, we train the models with local cross entropy loss and do early stopping based on the BLEU score on the dev set. For the inference network, we train the model to minimize the energy (Eq.~\ref{eq:objective}) and do early stopping based on the energy on the dev set.

\subsection{Predicting Target Sequence Lengths}
\label{sec:lengths}

Non-autoregressive models  
often need a target sequence length in advance~\citep{lee-etal-2018-deterministic}. 
We report results both with oracle lengths and 
with a simple method of predicting it. 
We follow \citet{ghazvininejad-etal-2019-mask} in predicting the length of the translation using a representation of the source sequence from the encoder. 
The length loss is added to the cross-entropy loss for the target sequence. During decoding, we select the top $k=3$ length candidates with the highest probabilities, 
decode with the different lengths in parallel, and return the translation with the highest average of log probabilities of its tokens.

\section{Results}

\begin{table}[t]
\centering
\small
\begin{tabular}{c c  c c c  c }
    \toprule
    & \multicolumn{2}{c}{IWSLT14 DE-EN} & & \multicolumn{2}{c}{WMT16 RO-EN} \\
    \cline{2-3} \cline{5-6}
    \noalign{\smallskip}
    & \multicolumn{2}{c}{\# iterations} & & \multicolumn{2}{c}{\# iterations} \\ 
    \cline{2-3} \cline{5-6}
    \noalign{\smallskip}
    & 1 &  10 & & 1 &  10  \\
    \midrule
 CMLM  & 28.11 & 33.39 & & 28.20 & 33.31\\
 \name  &  31.99 & 33.17 &  & 33.16 & 34.04\\
\bottomrule
\end{tabular}
\caption{Test BLEU scores of non-autoregressive models using no refinement (\# iterations = 1) and using refinement (\# iterations = 10). Note that the \# iterations = 1 results are purely non-autoregressive. \name uses a CMLM as the inference network architecture 
and the transformer AR energy. The length beam size is 5 for CMLM and 3 for \name. 
}
\label{tab:refinements}
\end{table}

\begin{table}[t]
\small
\setlength{\tabcolsep}{3pt}
\begin{center}
\begin{tabular}{l l  c   c }
\toprule
 &  & \multicolumn{1}{c}{IWSLT14} & \multicolumn{1}{c}{WMT16} \\
 &  & {DE-EN} & {RO-EN} \\ 
\midrule
 \multicolumn{2}{l}{\textbf{Autoregressive} (Transformer)} \\
\midrule
\multicolumn{1}{l}{} & Greedy Decoding  &  33.00 &  33.33 \\  
\multicolumn{1}{l}{} & Beam Search  & 34.11  & 34.07  \\ 
\midrule
 
\multicolumn{2}{l}{\textbf{Non-autoregressive}} \\
\midrule
& \makecell[l]{Iterative Refinement \\ \quad \citep{lee-etal-2018-deterministic}} & -& 25.73{{\makebox[0pt][l]{$^\dagger$}}}\\
& NAT with Fertility \citep{gu2018non} & - & 29.06{{\makebox[0pt][l]{$^\dagger$}}} \\
& CTC~\citep{libovicky-helcl-2018-end} & - & 24.71{{\makebox[0pt][l]{$^\dagger$}}} \\
& FlowSeq~\citep{ma-etal-2019-flowseq}  & 27.55{{\makebox[0pt][l]{$^\dagger$}}} & 30.44{{\makebox[0pt][l]{$^\dagger$}}} \\

\multicolumn{1}{l}{} & \makecell[l]{CMLM \\ \quad \citep{ghazvininejad-etal-2019-mask}}  & 28.25 &  28.20{{\makebox[0pt][l]{$^\dagger$}}} \\ 
\multicolumn{1}{l}{} &  \makecell[l]{Bag-of-ngrams-based loss \\ \quad \citep{shao2019minimizing}} & - & 29.29{{\makebox[0pt][l]{$^\dagger$}}}   \\
\multicolumn{1}{l}{} & \makecell[l]{AXE CMLM \\ \quad \citep{ghazvininejad2020aligned}} & - & 31.54{{\makebox[0pt][l]{$^\dagger$}}}  \\ 
\multicolumn{1}{l}{} & \makecell[l]{Imputer-based model\\ \quad \citep{saharia2020nonautoregressive}} & - & 31.7{{\makebox[0pt][l]{$^\dagger$}}}   \\ 
\multicolumn{1}{l}{} & {ENGINE (ours)}  & \textbf{31.99} & \textbf{33.16}\\
\bottomrule
\end{tabular}
\caption{BLEU scores on two datasets for several non-autoregressive methods. The inference network architecture is the CMLM. 
For methods that permit multiple refinement iterations (CMLM, AXE CMLM, ENGINE), one decoding iteration is used (meaning the methods are purely non-autoregressive). 
$^\dagger$Results are from the corresponding papers.  
\label{tab:finalInfnet}
}
\end{center}
\end{table}

\paragraph{Effect of choices for $\oone$ and $\otwo$.} 

Table~\ref{tab:argmax} compares various choices for the operations $\oone$ and $\otwo$. 
For subsequent experiments, we choose the setting that feeds the whole distribution into the energy function ($\oone$ = SX) and computes the loss with straight-through ($\otwo$ = ST). 
Using Gumbel noise in $\otwo$ has only minimal effect, and rarely helps. Using ST instead also speeds up training by avoiding the noise sampling step. 

\paragraph{Training with distilled outputs vs. training with energy.}
We compared training non-autoregressive models using the references, distilled outputs, and as inference networks on both datasets. 
Table~\ref{tab:cropus} in the Appendix shows the results when using BiLSTM inference networks and seq2seq AR energies. 
The inference networks improve over training with the references by 11.27 BLEU on DE-EN and 12.22 BLEU on RO-EN.  
In addition, inference networks consistently improve over non-autoregressive networks trained on the distilled outputs.

\paragraph{Impact of refinement iterations.}
\citet{ghazvininejad-etal-2019-mask} show improvements with multiple refinement iterations.  Table~\ref{tab:refinements} shows refinement results of CMLM and \name. Both improve with multiple iterations, though the improvement is much larger with CMLM. However, even with 10 iterations, \name is comparable to CMLM on DE-EN and outperforms it on RO-EN.

\paragraph{Comparison to other NAT models.}
Table~\ref{tab:finalInfnet} shows 1-iteration  results on two datasets. To the best of our knowledge, \name achieves state-of-the-art NAT performance: 31.99 on IWSLT14 DE-EN and 33.16 on WMT16 RO-EN. In addition, \name achieves comparable performance with the autoregressive NMT model.

\section{Conclusion}
We proposed a new method to train non-autoregressive neural machine translation systems via minimizing pretrained energy functions with inference networks. In the future, we seek to expand upon energy-based translation using our method.

\section*{Acknowledgments}

We would like to thank Graham Neubig for helpful discussions and the reviewers for insightful comments. 
This research was supported in part by an Amazon Research Award to K.~Gimpel.

\bibliography{anthology,acl2020}
\bibliographystyle{acl_natbib}

\appendix
\section{Appendix}

\subsection{Training with distilled outputs vs. training with energy.}

In order to compare \name with training on distilled outputs, 
we train BiLSTM models in three ways: ``baseline'' which is trained with the human-written reference translations, ``distill'' which is trained with the distilled outputs (generated using the autoregressive models), and  ``\name'', our method which trains the BiLSTM as an inference network to minimize the pretrained seq2seq autoregressive energy. Oracle lengths are used for decoding.  Table~\ref{tab:cropus} shows test results for both datasets, showing significant gains of \name over the baseline and distill methods. Although the results shown here are lower than the transformer results, the trend is clearly indicated.

\begin{table}[!th]
\setlength{\tabcolsep}{3pt}
\centering
\small
\begin{tabular}{ c | c c | c c}
    \toprule
      &  \multicolumn{2}{c|}{IWSLT14 DE-EN} & \multicolumn{2}{c}{WMT16 RO-EN} \\ 
      & {Energy ($\downarrow$) } &  {BLEU ($\uparrow$)} &  {Energy ($\downarrow$)} &  {BLEU ($\uparrow$)}\\
    \midrule
 baseline  & 153.54 & 8.28 & 175.94 & 9.47 \\
distill  & 112.36  & 14.58 & 205.71 & 5.76\\ 
\name & 51.98 & 19.55 & 64.03 & 21.69 \\ 
\bottomrule
\end{tabular}
\caption{Test results of non-autoregressive models when training with the references (``baseline''), distilled outputs (``distill''), and energy (``\name''). Oracle lengths are used for decoding. Here, \name uses BiLSTM inference networks and pretrained seq2seq AR energies.  \name  outperforms training on both the references and a pseudocorpus. 
}
\label{tab:cropus}
\end{table}

\subsection{Analysis of Translation Results}

\begin{table*}[th]
\setlength{\tabcolsep}{3pt}
\small
\centering
\begin{tabular}{|ll|}
	\hline
	\textbf{Source:} & \\ 
	\multicolumn{2}{|p{1.9\columnwidth}|}{
    seful onu a solicitat din nou tuturor partilor , inclusiv consiliului de securitate onu divizat sa se unifice si sa sustina negocierile pentru a gasi o solutie politica .}\\
	\textbf{Reference} : &\\ 
	\multicolumn{2}{|p{1.9\columnwidth}|}{the u.n. chief again urged all parties , including the divided u.n. security council , to unite and support inclusive negotiations to find a political solution .} \\
	\textbf{CMLM} :&\\ 
	\multicolumn{2}{|p{1.9\columnwidth}|}{the un chief again again urged all parties , including the divided un security council to unify and support negotiations in order to find a political solution .} \\
	\textbf{\name} :  &\\ 
	\multicolumn{2}{|p{1.9\columnwidth}|}{the un chief has again urged all parties , including the divided un security council to unify and support negotiations in order to find a political solution .} \\
    \hline
    \end{tabular}

\begin{tabular}{|ll|}
	\hline
	\textbf{Source:} & \\ 
	\multicolumn{2}{|p{1.9\columnwidth}|}{adevarul este ca a rupt o racheta atunci cand a pierdut din cauza ca a acuzat crampe in us , insa nu este primul jucator care rupe o racheta din frustrare fata de el insusi si il cunosc pe thanasi suficient de bine incat sa stiu ca nu s @-@ ar mandri cu asta .
    }\\
	\textbf{Reference} : &\\ 
	\multicolumn{2}{|p{1.9\columnwidth}|}{he did break a racquet when he lost when he cramped in the us , but he \&apos;s not the first player to break a racquet out of frustration with himself , and i know thanasi well enough to know he wouldn \&apos;t be proud of that .} \\
	\textbf{CMLM} :&\\ 
	\multicolumn{2}{|p{1.9\columnwidth}|}{the truth is that it has broken a rocket when it lost because accused crcrpe in the us , but it is not the first player to break rocket rocket rocket frustration frustration himself himself and i know thanthanasi enough enough know know he would not be proud of that .} \\
	\textbf{\name} :  &\\ 
	\multicolumn{2}{|p{1.9\columnwidth}|}{the truth is that it broke a rocket when it lost because he accused crpe in the us , but it is not the first player to break a rocket from frustration with himself and i know thanasi well well enough to know he would not be proud of it .} \\
    \hline
    \end{tabular}
    
    \begin{tabular}{|ll|}
	\hline
	\textbf{Source:} & \\ 
	\multicolumn{2}{|p{1.9\columnwidth}|}{realizatorii studiului mai transmit ca \&quot; romanii simt nevoie de ceva mai multa aventura in viata lor ( 24 \% ) , urmat de afectiune ( 21 \% ) , bani ( 21 \% ) , siguranta ( 20 \% ) , nou ( 19 \% ) , sex ( 19 \% ) , respect 18 \% , incredere 17 \% , placere 17 \% , conectare 17 \% , cunoastere 16 \% , protectie 14 \% , importanta 14 \% , invatare 12 \% , libertate 11 \% , autocunoastere 10 \% si control 7 \% \&quot; .
    }\\
	\textbf{Reference} : &\\ 
	\multicolumn{2}{|p{1.9\columnwidth}|}{the study \&apos;s conductors transmit that \&quot; romanians feel the need for a little more adventure in their lives ( 24 \% ) , followed by affection ( 21 \% ) , money ( 21 \% ) , safety ( 20 \% ) , new things ( 19 \% ) , sex ( 19 \% ) respect 18 \% , confidence 17 \% , pleasure 17 \% , connection 17 \% , knowledge 16 \% , protection 14 \% , importance 14 \% , learning 12 \% , freedom 11 \% , self @-@ awareness 10 \% and control 7 \% . \&quot; } \\
	\textbf{CMLM} :&\\ 
	\multicolumn{2}{|p{1.9\columnwidth}|}{survey survey makers say that \&apos; romanians romanians some something adventadventure ure their lives 24 24 \% ) followed followed by \% \% \% \% \% , ( 21 \% \% ), safety ( \% \% \% ), new19\% \% ), ), 19 \% \% \% ), respect 18 \% \% \% \% \% \% \% \% , , \% \% \% \% \% \% \% , , \% , 14 \% , 12 \% \% } \\
	\textbf{\name} :  & \\ 
	\multicolumn{2}{|p{1.9\columnwidth}|}{realisation of the survey say that \&apos; romanians feel a slightly more adventure in their lives ( 24 \% ) followed by aff\% ( 21 \% ) , money ( 21 \% ), safety ( 20 \% ) , new 19 \% ) , sex ( 19 \% ) , respect 18 \% , confidence 17 \% , 17 \% , connecting 17 \% , knowledge \% \% , 14 \% , 14 \% , 12 \% \%} \\
    \hline
    \end{tabular}
    \caption{Examples of translation outputs from \name and CMLM on WMT16 RO-EN without refinement iterations.}
    \label{tab:translations}
\end{table*}

In Table~\ref{tab:translations}, we present randomly chosen translation outputs from WMT16 RO-EN. For each Romanian sentence, we show the reference from the dataset, the translation from CMLM, and the translation from \name. We could observe that without the refinement iterations, CMLM could performs well for shorter source sentences. However, it still prefers generating repeated tokens. \name, on the other hand, could generates much better translations with fewer repeated tokens.

\end{document}